\documentclass{bmvc2k}

\title{Domain-Aware Augmentations for Unsupervised Online General Continual Learning}

\addauthor{Nicolas Michel}{nicolas.michel@esiee.fr}{1}
\addauthor{Romain Negrel}{romain.negrel@esiee.fr}{1}
\addauthor{Giovanni Chierchia}{giovanni.chierchia@esiee.fr}{1}
\addauthor{Jean-François Bercher}{jf.bercher@esiee.fr}{1}

\addinstitution{
 Univ Gustave Eiffel \\
 CNRS \\
 LIGM \\
 F-77454 Marne-la-Vallée, France
}

\runninghead{Michel, Negrel, Chierchia, Bercher}{Domain-Aware Augmentations}

\DeclareMathOperator{\MemoryUpdate}{MemoryUpdate}
\DeclareMathOperator{\Retrieve}{Retrieve}

\DeclareMathOperator{\Aug}{Aug}
\DeclareMathOperator{\DAA}{DAA}

\DeclareMathOperator{\DAM}{DAM}
\DeclareMathOperator{\DAS}{DAS}

\usepackage{amsmath}
\usepackage{amssymb}
\usepackage{mathtools}
\usepackage{amsthm}
\usepackage{multirow}

\usepackage[capitalize,noabbrev]{cleveref}

\theoremstyle{plain}

\theoremstyle{definition}

\theoremstyle{remark}

\let\svthefootnote\thefootnote
\newcommand\freefootnote[1]{%
  \let\thefootnote\relax%
  \footnotetext{#1}%
  \let\thefootnote\svthefootnote%
}

\usepackage{algorithm}
\usepackage[noend]{algpseudocode}

\begin{document}

\maketitle

\begin{abstract}
    Continual Learning has been challenging, especially when dealing with unsupervised scenarios such as Unsupervised Online General Continual Learning (UOGCL), where the learning agent has no prior knowledge of class boundaries or task change information. While previous research has focused on reducing forgetting in supervised setups, recent studies have shown that self-supervised learners are more resilient to forgetting. This paper proposes a novel approach that enhances memory usage for contrastive learning in UOGCL by defining and using stream-dependent data augmentations together with some implementation tricks. Our proposed method is simple yet effective, achieves state-of-the-art results compared to other unsupervised approaches in all considered setups, and reduces the gap between supervised and unsupervised continual learning. Our domain-aware augmentation procedure can be adapted to other replay-based methods, making it a promising strategy for continual learning.
\end{abstract}

\section{Introduction}
\label{intro}
\freefootnote{This work has received support from Agence Nationale de la Recherche (ANR) for the project APY, with reference ANR-20-CE38-0011-02. This work was granted access to the HPC resources of IDRIS under the allocation 2022-AD011012603 made by GENCI.}
Continual Learning (CL) is the ability to learn from a continuously evolving stream of data while accommodating shifts in distribution over time. Recent years have witnessed numerous attempts to simulate such an environment for image classification, including domain and class-incremental learning scenarios \cite{hsu_re-evaluating_2019}. While much of the prior research has been focused on a fully supervised scenario that assumes specific prior knowledge, unsupervised CL methods operate under more challenging circumstances where there is no task boundary or the total number of classes available. This work focuses on a more realistic learning scenario where only one pass over non-iid, unlabeled data is allowed without prior task knowledge, task change information, or known number of classes during training. This setup is known as Unsupervised Online General Continual Learning (UOGCL) \cite{buzzega_dark_2020} and only a handful of approaches have been designed to address it. STAM \cite{smith_unsupervised_2021} employs a patch-based online clustering with novelty detection and expandable memory. SCALE \cite{yu_scale_2023} leverages a pseudo-labeled contrastive loss and knowledge distillation with a fixed memory to learn data representation. By design, both STAM and SCALE strongly focus on reducing forgetting.

Although forgetting is widely recognized as the main issue in CL environments, self-supervised learners have been found to be exceptionally resilient to forgetting compared to cross-entropy trained models \cite{madaan_lump_2022,fini_cassle_2022,davari_probing_2022}. Additionally, several studies demonstrate that replay-based methods can easily take advantage of memory data more efficiently. One way is to use implementation tricks for reviewing memory data \cite{mai_online_2021,liang_new_2023}, and another is to train for multiple iterations for each batch \cite{buzzega_rethinking_2020,mai_online_2021}. Similarly, some methods have obtained state-of-the-art results while training using memory data only \cite{vedaldi_gdumb_2020, michel_contrastive_2022}. Previous observations indicate that replay-based self-supervised learners might not need anti-forgetting mechanisms to cope with UOGCL. Rather, a promising strategy would be to learn more efficiently from memory data.

This paper focuses on replay-based methods showing the best performances in online CL. We introduce a novel replay-based method that improves memory utilization with contrastive loss by combining stream-dependent data augmentations with implementation tricks for UOGCL. Despite its simplicity, our method performs better than other unsupervised methods in all evaluation setups. Additionally, the proposed Domain-Aware Augmentation procedure could easily be integrated into other replay-based approaches with minor adaptations to improve their performance as well.

The paper is structured as follows: Section \ref{sec:related} presents related work. Section \ref{sec:method_def} describes the training procedure, the strategy used to improve memory usage, and our new Domain-Aware Augmentation framework for replay-based methods. Section \ref{sec:exps} relates our experiments and eventually, section \ref{sec:conclusion} concludes the paper.

\begin{figure}[ht!]
\vskip -0.2in
\begin{center}
\centerline{\includegraphics[width=\columnwidth]{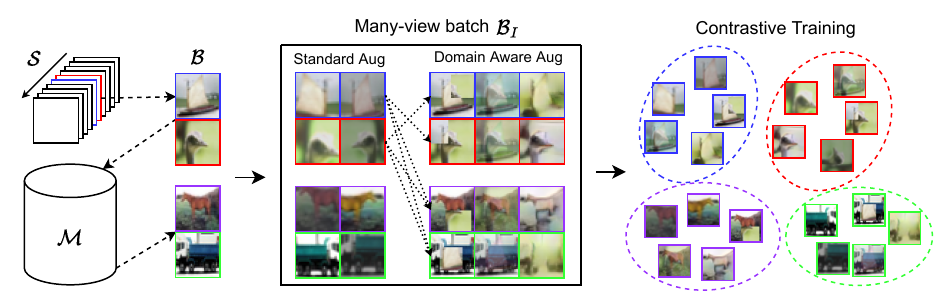}}
\vskip -0.15in
\caption{Overview of the Domain-Aware Augmentation procedure. From left to right, unlabeled images are sampled from stream $\mathcal{S}$ and memory $\mathcal{M}$ to create the incoming batch $\mathcal{B}$. This batch is augmented to obtain a many-view batch $\mathcal{B}_I$. Here $\mathcal{B}_I$ is composed of 2 standard augmentations and 3 $\DAA$. Images from $\mathcal{S}$ are used to create DAA for every image in $\mathcal{B}$. The model then learns image representation by minimizing the contrastive loss defined in eq. \ref{eq:mvcont}. Best viewed in color.}
\label{fig:daa}
\end{center}
\vskip -0.3in
\end{figure}

\section{Related work}
This section defines learning strategies related to the work presented here.
\label{sec:related}
\subsection{Online General Continual Learning}
In the following, we define the considered CL setups.

\textbf{Online Continual Learning (OCL)} addresses the problem of learning from a continuous stream of data. Formally, we consider a sequential learning setup with a sequence $\{\mathcal{T}_1,\cdots,\mathcal{T}_K\}$ of $K$ tasks, and $\mathcal{D}_k=(X_k, Y_k)$ the corresponding data-label pairs. In CL, we often assume that for any value $k_1,k_2 \in \{1,\cdots,K\}$, if $k_1\neq k_2$ then we have $Y_{k_1}\cap Y_{k_2}=\emptyset$ and the number of classes in each task is the same. Contrary to standard CL, in OCL only one pass over the data is allowed. This setup has been studied mostly in a fully supervised scenario \cite{aljundi_gradient_2019, aljundi_online_2019, mai_online_2021, rolnick_experience_2019, mai_supervised_2021,vedaldi_gdumb_2020, guo_online_2022}.

\textbf{Online General Continual Learning (OGCL)} imposes further constraints on the already challenging task of Continual Learning. In this setup, the learning model is not provided with any prior information about the training environment, including task-ids, task boundaries, number of classes per task, total classes, and tasks. While previous research efforts have mainly focused on developing methods for the supervised OGCL scenario \cite{mai_online_2021, mai_supervised_2021, rolnick_experience_2019, aljundi_online_2019}, only a limited number of approaches have been proposed for the unsupervised case. Among them, STAM \cite{smith_unsupervised_2021} and SCALE \cite{yu_scale_2023} were recently introduced to address the challenges of Unsupervised Online General Continual Learning (UOGCL).

\textbf{Replay based methods}.
In replay-based CL methods, a memory buffer stores a subset of the past training samples. As the learning model encounters a new batch of data from the data stream, a corresponding batch is retrieved from the memory, and the model is trained on the combined set of both the stream and memory batches. During the interval between two consecutive stream batches, the memory is updated with the most recent data from the stream batch. Replay-based methods have been widely developed in CL \cite{aljundi_online_2019,mai_online_2021,michel_contrastive_2022,buzzega_dark_2020,rolnick_experience_2019,guo_online_2022,yu_scale_2023}.

\textbf{Contrastive Learning}.
\label{sec:contrastive}
Contrastive Learning has become a widely used technique to learn image representations \cite{chen_simple_2020,He_2020_CVPR}. The essential principle underlying this approach is to train a model that maps similar data samples (referred to as positives) into closer proximity in a feature space while pushing dissimilar data samples (referred to as negatives) away from each other. In situations where labeled data is not available, augmentations of the same image are treated as positives, while all other images are considered negatives. Mai et al. \cite{mai_supervised_2021} used contrastive learning in the supervised scenario, and unsupervised contrastive learning was used recently for UOGCL by Yu et al. \cite{yu_scale_2023}. It was also adapted to a semi-supervised scenario by Michel et al. \cite{michel_contrastive_2022}.

\textbf{Average Accuracy}.
Average Accuracy (AA) is the standard metric used in Continual Learning. It measures the overall accuracy of a model at each task, averaged across all tasks learned up to that point. In this paper, we focus exclusively on the final Average Accuracy as our evaluation metric \cite{hsu_re-evaluating_2019, mai_online_2021}, which is equivalent to the accuracy of the model at the end of training. By using only the final Average Accuracy, we can get a clear picture of how well a model has performed over the entire learning process. This allows a fair comparison of different approaches to Continual Learning and provides a consistent performance measure.

\section{Method Definition}
\label{sec:method_def}
In this section, we define our method. First, we describe our training procedure. Second, we discuss the impact of key hyper-parameters, and last, we introduce a new augmentation strategy for continual learning.

\subsection{Training procedure}
In the following, we define the training procedure of our method.

\textbf{Many-view batch}.
We propose an extension to the multi-view batch concept as described by Khosla et al. \cite{khosla_supervised_2020} that involves using more than two augmentations. Specifically, we define the many-view batch as the union of $p$ augmentations for a batch $\mathcal{B}$ such that $\mathcal{B}_I=\mathcal{B}\bigcup_{i=1}^{p}\Aug(\mathcal{B})$, where $p$ is the number of augmentations and $I$ represents the indices over $\mathcal{B}_I$. To train the model on many-view batches, we adapt the SupCon loss for unsupervised scenarios by treating every augmentation of the same input image as having the same label. We formulate this approach as minimizing the Multi-View Contrastive (MVCont) loss, defined as follows:
\begin{equation}
\mathcal{L}_{MVCont}(\mathcal{B}_I, \theta) = \displaystyle{-\sum_{i\in I}\frac{1}{\vert P(i) \vert}\sum_{p\in P(i)}\log\frac{e^{z_i\cdot z_p/\tau}}{\displaystyle\sum_{a \in I \backslash \{i\}}e^{z_i\cdot z_a/\tau}}}
\label{eq:mvcont}
\end{equation}
Here, $P(i) = \{j \in I \backslash \{i\}\ \vert\ y_j=y_i\}$ represents the set of images having the same input source as input $i$, $Z_I=\{z_i\}_{i \in I}$, $f_\theta$ denotes the learnable model with parameters $\theta$, and $z_i=f_\theta(x_i)$ represents the feature vector of the input image $x_i$.

\textbf{Experience Replay with Contrastive Learning}.
We propose to combine Experience Replay (ER) \cite{rolnick_experience_2019} with unsupervised contrastive learning on a many-view batch by minimizing $\mathcal{L}_{MVCont}$ defined in equation \ref{eq:mvcont}.  Similar to ER, we mitigate forgetting by using a fixed sized memory that is filled following a reservoir sampling strategy \cite{vitter_random_1985} and a random retrieval. The overall training procedure is detailed in Algorithm \ref{algo:replay_based_methods}.
\begin{algorithm}
    \begin{algorithmic}
        \State\textbf{Input:}\ Data stream $\mathcal{S}$; Augmentation procedure $\Aug(.)$; Model $f_\theta(.)$
        \State\textbf{Output:}\ Model $f_\theta$; Memory $\mathcal{M}$
        \State $\mathcal{M} \gets$ \{\} \Comment{Initialize memory}
        \For{$\mathcal{B}_s \in \mathcal{S}$}  \Comment{Data stream}
            \For{$q$ iterations} \Comment{Memory iterations}
                \State $\mathcal{B}_m \gets \Retrieve(\mathcal{M})$ \Comment{Retrieve data from memory}
                \State $\mathcal{B} \gets \mathcal{B}_s \cup \mathcal{B}_m$ \Comment{Combined Batch}
                \State $\mathcal{B_I} \gets \mathcal{B}\bigcup_{i=1}^p \Aug(\mathcal{B})$ \Comment{Many-view batch}
                \State $\theta \gets SGD(\mathcal{L}_{MVCont}(f_\theta(\mathcal{B_I}), \theta))$ \Comment{Loss defined in \ref{eq:mvcont}}
                \EndFor
            \State $\mathcal{M} \gets \ \MemoryUpdate(\mathcal{B}_s, \mathcal{M})$
        \EndFor
        \State\textbf{return:}\ $\theta$;\ $\mathcal{M}$
    \end{algorithmic}
    \caption{Proposed Training Procedure \label{algo:replay_based_methods}}
\end{algorithm}

\subsection{Improving memory usage}
\label{sec:mem_data}
In the following, we discuss several strategies to improve memory usage in the training procedure defined in Algorithm \ref{algo:replay_based_methods}. Experimental results regarding such tricks are presented in Table \ref{tab:mem_params_impact}.

\textbf{Larger Memory batch size $\vert\mathcal{B}_m\vert$}.
One common hyper-parameter impacting the performance of replay-based methods is the memory batch size $\vert\mathcal{B}_m\vert$, the amount of data retrieved from memory when encountering a new stream batch. As the size of $\vert\mathcal{B}_m\vert$ increases, the model will be exposed to memory data more frequently, which can lead to overfitting. However, in UOGCL, we found that increasing $\vert\mathcal{B}_m\vert$ results in steadily increasing performances.

\textbf{More Memory Iterations $q$}.
In Algorithm \ref{algo:replay_based_methods}, $q$ represents the number of memory iterations, indicating how often the model will be exposed to memory data during training. As $q$ increases, the model will have more opportunities to learn from the memory data and potentially improve its performance on the task at hand. This technique has been applied in previous works \cite{aljundi_online_2019} with supervised methods with the risk of overfitting to the current task. In UOGCL, we observe little overfitting.

\textbf{More augmentations}.
\label{sec:more_augs}
Using more data augmentation can improve the learning process in online continual learning scenarios. It helps the model learn better by enabling it to see the same data from different perspectives, recognize patterns, and generalize. Augmentation also generates new training samples from existing ones, making the model adaptable to evolving data distributions. In that sense, increasing the value of $p$, the number of views in the many-view batch can similarly increase performances. However, standard augmentations like random crop and color-jitter are limited as they do not use external information. For example, a random crop augmentation only has a limited number of crops and throughout training, the model is likely to be trained on every variation of augmented memory data, encouraging overfitting. This phenomenon is exacerbated when using multiple memory iterations. Therefore, more sophisticated augmentations are presented in section \ref{sec:daa}.


\subsection{Domain Aware Augmentations (DAA)}
\label{sec:daa}
As introduced in section \ref{sec:more_augs}, traditional data augmentation can be limited for replay methods. This section proposes a framework for stronger domain-aware augmentations that leverages stream information. This allows the model to view memory data through an unlimited amount of perspectives along training.


\textbf{DAA framework}
\label{sec:daa_framework}
We define a DAA as an augmentation that combines an input image $x_i$ with a domain-related image $x_d$, resulting in an augmented version of $x_i$ denoted as $x_a=\DAA(x_i, x_d)$ via the $\DAA$ procedure. In replay-based approaches, $x_i$ comes from the current batch $\mathcal{B}$, while $x_d$ comes from the stream.


\textbf{Domain-Aware Mixup (DAM)}.
Mixup has been introduced in 2018 \cite{zhang_mixup_2018} in the supervised scenario as a new augmentation technique that linearly interpolates between two data-label pairs. Recently, mixup has been adapted to the CL setting \cite{douillard_dytox_2022,madaan_lump_2022}. Notably, in LUMP \cite{madaan_lump_2022}, Madaan et al. introduced mixup strategies between memory and stream images to create new images for replay-based unsupervised CL. For $x_i \in \mathcal{M}$ from memory and $x_d \in \mathcal{S}$ from stream the author trained a model on $x_a = \lambda\cdot x_i + (1- \lambda)\cdot x_d$. Notably, the obtained images are considered as entirely new images. In this work, we define DAM by constructing augmented images $x_a=\lambda\cdot x_i + (1-\lambda)\cdot x_d$, however, mixup-generated images are used as views of the original image. Additionally, we use $\lambda \sim \mathcal{U}(0.5,1)$, $x_i \in \mathcal{B}$, $x_d \in \mathcal{S}$ and $x_a=\DAM(x_i, x_d)$. The interpolation factor is set such that the augmented image $x_a$ has at least half of its information coming from the input image $x_i$. This strategy is inspired by the SMOTE \cite{chawla_smote_2002} oversampling strategy.

\textbf{Domain-Aware CutMix (DAC)}. CutMix is another augmentation technique \cite{yun_cutmix_2019}, which bears similarities with mixup. Likewise to the DAM adaptation we consider $x_i \in \mathcal{B}$ and $x_d \in \mathcal{S}$ to create $x_a$, a new view of $x_i$ such that $x_a =M\odot x_i + (\textbf{1}-M)\odot x_d$
with $M\in\{0,1\}^{W\times H}$ a binary mask where $W$ and $H$ are the width and the height of the image. \textbf{1} is a binary mask filled with ones, $\odot$ is the Hadamard product and $\lambda \sim \mathcal{U}(0.5,1)$. The binary mask is constructed according to the bounding box coordinates $B = (r_x, r_y, r_w, r_h)$ which correspond to the region to crop from $x_d$ and integrate into $x_i$. Following the work proposed by \cite{yun_cutmix_2019} we sample the bounding box for a given $\lambda$ value according to:
\begin{equation}
    \begin{aligned}
        r_x \sim \mathcal{U}(0,W),&\ \  r_w = W\sqrt{1-\lambda} \\
        r_h \sim \mathcal{U}(0,H),&\ \  r_h = H\sqrt{1-\lambda}
    \end{aligned}
\end{equation} 
As with DAM we use $\lambda\ge0.5$ to ensure that a significant part of the original image is present in the augmented version.

\textbf{Domain-Aware Style (DAS)}. Style transfer is the transfer of non-semantic visual information from one image $x_d$ to another image $x_i$ to create the resulting image $x_a$, with content from $x_i$ and style from $x_d$. The original style transfer method proposed by \cite{gatys_image_2016} relies on a slow optimization process which cannot reasonably be applied as a data augmentation procedure. \cite{huang_arbitrary_2017} proposed a method based on instance normalization that can compute and transfer any style from any image efficiently, but has to be pre-trained beforehand. A model pre-trained on MS-COCO \cite{lin_microsoft_2015} is used to transfer the style from $x_d \in \mathcal{S}$ to $x_i\in \mathcal{B}$. The obtained image is considered as another view of $x_i$ such that $x_a=\DAS(x_i, x_d)$.

\section{Experiments}
\label{sec:exps}
In this section, we first describe our setup: evaluation protocol, datasets used, baseline methods considered for comparisons, and implementation details; before presenting our experimental results.

\subsection{Evaluation Protocol}
\label{sec:eval_protocol}
Since we focus on UOGCL, the training procedure defined in Algorithm \ref{algo:replay_based_methods} outputs a trained encoder $f_\theta(.)$ and a subset of images $\mathcal{M}$. An extra transfer-learning step is required for classification. For a fair comparison, we use only the images stored in memory $\mathcal{M}$ at the end of training for transfer learning. This is equivalent to adding an extra step for labeling memory after training. As it in common in representation learning \cite{mai_online_2021, chen_simple_2020, fini_cassle_2022} we consider the trained model $f_\theta(.)$ as being the succession of a feature extractor $h_{\theta_r}(.)$ and a projection head $g_{\theta_p}(.)$ such that $f_\theta(.)=g_{\theta_p}(h_{\theta_r}(.))$. For the transfer learning step, the representations obtained from $h_{\theta_r}(.)$ are used, as described in Algorithm \ref{algo:eval_proto}.
\begin{algorithm}
\begin{algorithmic}
\State\textbf{Input:}\ Data stream $\mathcal{S}$; Memory $\mathcal{M}$; Augmentation procedure $\Aug(.)$; Feature extractor $h_{\theta_r}(.)$; Projection head $g_{\theta_p}(.)$; Nearest Class Mean classifier $\phi_\omega(.)$
\State\textbf{Output:}\ End-to-end classifier $\phi_\omega(h_{\theta_r}(.))$
\State\texttt{Training Phase:}
\State $\theta_r$, $\mathcal{M} \gets$ Train$(\mathcal{B}_s, Aug(.), f_\theta(.))$ \Comment{Train as in Algorithm \ref{algo:replay_based_methods} with $f_\theta(.)=g_{\theta_p}(h_{\theta_r}(.))$}
\State\texttt{Testing Phase:}
\State $R \gets$ $h_{\theta_r}(\mathcal{M})$
\State $\omega \gets$ TrainNCM($\omega$, $R$) \Comment{Train a Nearest Class Mean classifier on representations.}
\State \textbf{return:}\ $\omega$;\ $\theta_r$

\end{algorithmic}
\caption{Proposed Evaluation procedure\label{algo:eval_proto}}
\end{algorithm}

\begin{table}[ht!]
\centering
\setlength{\tabcolsep}{4pt}
\resizebox{0.7\textwidth}{!}{
    \begin{tabular}{c | c | c c | c c | c c }
        \multicolumn{2}{c}{} & \multicolumn{2}{c}{\rule{0pt}{2ex}CIFAR10} & \multicolumn{2}{c}{CIFAR100} & \multicolumn{2}{c}{Tiny IN} \\
        \hline
        \hline
                                    & 10 & \multicolumn{2}{c|}{\rule{0pt}{2.3ex}34.7±1.8} & \multicolumn{2}{c|}{11.3±0.4}     & \multicolumn{2}{c}{8.8±0.04} \\
        Memory                      & 20 & \multicolumn{2}{c|}{36.3±2.7}                 & \multicolumn{2}{c|}{11.8±1.0}      & \multicolumn{2}{c}{10.1±0.2} \\
        batch size                  & 50 & \multicolumn{2}{c|}{41.1±2.0}                & \multicolumn{2}{c|}{16.8±1.0}       & \multicolumn{2}{c}{13.2±0.5} \\
        $\vert\mathcal{B}_m\vert$   & 100 & \multicolumn{2}{c|}{42.9±0.1}               & \multicolumn{2}{c|}{19.2±0.5}       & \multicolumn{2}{c}{15.2±0.3} \\
                                    & 200 & \multicolumn{2}{c|}{\textbf{43.2±2.3}}      & \multicolumn{2}{c|}{\textbf{21.2±0.9}} & \multicolumn{2}{c}{\textbf{16.7±0.5}} \\
        \hline
        \multicolumn{2}{c|}{} & \multicolumn{2}{c|}{\rule{0pt}{2ex}$\vert\mathcal{B}_m\vert=200$} & \multicolumn{2}{c|}{$\vert\mathcal{B}_m\vert=200$} & \multicolumn{2}{c}{$\vert\mathcal{B}_m\vert=200$} \\
        \hline
                    & 1 & \multicolumn{2}{c|}{43.2±2.3}             & \multicolumn{2}{c|}{21.2±0.9} & \multicolumn{2}{c}{16.7±0.5} \\
        Memory      & 2 & \multicolumn{2}{c|}{44.0±1.5}                    & \multicolumn{2}{c|}{23.1±0.2} & \multicolumn{2}{c}{17.2±0.6} \\
        iterations  & 3 & \multicolumn{2}{c|}{44.0±2.0}             & \multicolumn{2}{c|}{23.0±0.3} & \multicolumn{2}{c}{\textbf{18.3±0.3}} \\
        $q$         & 4 & \multicolumn{2}{c|}{\textbf{45.2±2.7}}    & \multicolumn{2}{c|}{23.8±0.4} & \multicolumn{2}{c}{17.6±0.2} \\
                    & 5 & \multicolumn{2}{c|}{42.6±1.9}             & \multicolumn{2}{c|}{\textbf{24.0±0.4}} & \multicolumn{2}{c}{18.1±0.5} \\
        \hline
        \multicolumn{2}{c|}{} & \rule{0pt}{2ex}$q=1$ & $q=4$ & $q=1$ & $q=4$ & $q=1$ & $q=4$ \\
        \hline
                    & 1  & \rule{0pt}{2ex}43.2±2.3  & \textbf{45.2±2.7}                 &  21.2±0.9    &  23.8±0.4         & 16.7±0.5     & 17.6±0.2 \\
        Number      & 2  & 44.4±0.5                 & 42.4±2.0    & 24.6±0.7            &  24.6±1.0         & 17.2±0.6     & 18.8±0.6 \\
        of          & 3  & 45.6±1.4                 & 41.8±5.0    & 25.7±0.4            &  25.9±0.6         & 18.0±0.4     & 18.7±0.4 \\
        views       & 4  & 45.3±1.7                 & 41.5±5.7    & 26.4±0.2            &  26.3±0.3         & 17.9±0.1     & 18.6±0.0 \\
        $p$         & 5  & 45.6±1.0                 & 39.0±6.1    & 26.7±0.3            &  \textbf{27.3±0.7}         & \textbf{18.2±0.4}     & \textbf{19.1±0.2} \\ 
                    & 6  & \textbf{45.7±1.0}        & 40.0±7.7    & \textbf{26.8±0.5}   &  26.8±0.1        & 18.1±0.4     & 18.5±0.9 \\
        \hline
    \end{tabular}
}
\vskip 0.15in
\caption{\label{tab:mem_params_impact} Impact of $\vert\mathcal{B}_m\vert$, $q$ and $p$ on the final AA (\%) for CIFAR10, CIFAR100 and Tiny ImageNet. The top part shows performances for $\vert\mathcal{B}_m\vert \in \lbrack10,200\rbrack$, $p=1$, $q=1$. The middle part shows performances for $q \in \lbrack1, 5\rbrack$, $\vert\mathcal{B}_m\vert=200$, $p=1$. The bottom part show performances for $p\in \lbrack1,6\rbrack$, $q\in \{1,5\}$, $\vert\mathcal{B}_m\vert=200$. The performances are obtained by following algorithm \ref{algo:eval_proto}. We use standard augmentations described in section \ref{sec:implem_details}. Each experiment is run 3 times and their average and standard deviation are displayed. The best results are displayed in bold.}
\end{table}

\subsection{Datasets}
We use variations of standard image classification datasets \cite{krizhevsky_learning_2009,le_tiny_2015} to build continual learning environments. The original datasets are split into several tasks of non-overlapping classes. Specifically, we experimented on split-CIFAR10, split-CIFAR100 and split-Tiny ImageNet. In this paper, we omitted the split- suffix for simplicity. \textbf{CIFAR10} contains 50,000 32x32 train images and 10,000 test images and is split into 5 tasks containing 2 classes each for a total of 10 distinct classes. \textbf{CIFAR100} contains 50,000 32x32 train images and 10,000 test images and is split into 10 tasks containing 10 classes each for a total of 100 distinct classes. \textbf{Tiny ImageNet} is a subset of the ILSVRC-
2012 classification dataset and contains 100,000 64x64 train images as well as 10,000 test images and is split into 20 tasks containing 10 classes each for a total of 200 distinct classes.
\subsection{Baselines}
In the following, we describe considered baselines. While proposing an unsupervised approach, we compare our method to supervised an unsupervised baselines to better demonstrate its efficiency. For methods using replay strategies, we add the suffix \textit{-ER} to the name and use reservoir sampling \cite{vitter_random_1985} for memory update and random retrieval. 
\textbf{fine-tuned}: Supervised lower bound corresponding to training using a cross entropy loss in a continual learning setup without precautions to avoid forgetting.\\
\textbf{offline}: Supervised upper bound. The model is trained without any CL specific constraints.\\
\textbf{Experience Replay} (ER) \cite{rolnick_experience_2019}: ER is a supervised memory based technique using reservoir sampling \cite{vitter_random_1985} for memory update and random retrieval. The model is trained using cross-entropy.\\
\textbf{Supervised Contrastive Replay} (SCR) \cite{mai_supervised_2021}: Replay-based method trained using the SupCon loss \cite{khosla_supervised_2020}. \\
\textbf{ER-ACE}~\cite{caccia_new_2022}: Replay based method using an Asymmetric Cross Entropy to overcome feature drift. \\
\textbf{GSA}~\cite{guo_dealing_2023}: Replay-based method dealing with cross-task class discrimination with a redefined loss objective using Gradient Self Adaptation. \\
\textbf{GDumb} \cite{vedaldi_gdumb_2020}: Simple method that stores data from the stream in memory, with the constraint of having a balanced class selection. At inference time, the model is trained offline on memory data. \\
\textbf{SimCLR-ER} \cite{chen_simple_2020}: Memory-based approach where the model is trained using the unsupervised contrastive loss of SimCLR. The memory management strategy is the same as the one used in ER.  \\
\textbf{BYOL-ER} \cite{grill_bootstrap_2020}: Memory-based approach where the model is trained using the loss defined in BYOL. The memory management strategy is the same as the one used in ER. \\
\textbf{SimSiam-ER} \cite{chen_exploring_2020}: Memory-based approach where the model is trained using the loss defined in SimSiam. The memory management strategy is the same as the one used in ER. \\
\textbf{LUMP} \cite{madaan_lump_2022}: Replay-based approach where every image in the batch is a mixup between memory and stream image. The model is trained using the unsupervised contrastive loss. Originally proposed in a non-online scenario, this method was adapted to the UOGCL.\\
\textbf{SCALE} \cite{yu_scale_2023}: Replay-based method using a pseudo-labeled contrastive loss. While very recent, the code is not available for this method and we had to report the available performances from the original paper.\\
\textbf{STAM} \cite{smith_unsupervised_2021}: A method designed for UOGCL using an expandable memory, patch-based clustering and novelty detection.

\subsection{Implementation details}
\label{sec:implem_details}
We train a ResNet-18 \cite{he_deep_2015} from scratch for every experiment. The projection layer for contrastive approaches is a MLP with 1 hidden layer of size 512, ReLU activation, and output size of 128. Memory batch size for replay-based methods is 200 and stream batch size for any method is 10. Our method uses an SGD optimizer with a fixed learning rate of 0.1. For all methods, a small hyperparameter search is conducted, and best parameters are kept for training. The search includes learning rate and optimizer. Temperature for contrastive losses is set to 0.07. For standard augmentations, we use random crop, colo jitter, random flip, and grayscale. Offline methods are trained for 50 epochs with the same optimizer, model, and augmentation procedure as other methods. Unsupervised methods are evaluated using NCM on memory data at the end of training following sec \ref{sec:eval_protocol}. For each experiment, the order of the labels for the training sequence is generated randomly.

    
\subsection{Results}
In what follows, we present our experimental results, highlighting the main figures and characteristics that demonstrate the interest and relevance of our approach.

\textbf{Scaling memory parameters.} Memory parameters described in \ref{sec:mem_data} can have a significant impact on performances. While expanding the amount of data retrieved from memory $|\mathcal{B}_m|$ continuously improves performances, it cannot exceed memory size. Similarly, we observe that increasing the amount of augmentation $p$ also results in an increase in performances for all datasets. However, larger values of memory iteration $q$ do not scale well for $p\geq 5$ while considerably increasing computation. Therefore, we set $q=1$ for our final method and scale with the number of augmentations rather than the number of iterations. However, experimenting with larger values of $q$ could lead to even higher performances.

\textbf{Final AA.} We report the final AA on table \ref{tab:avg_acc} for all methods. Our approach outperforms every other unsupervised method for UOGCL, on all considered setups. Notably, Ours - $(7,1,0,0,0)$, which corresponds to training with $(p,q)=(7,1)$ demonstrate that training with more augmentations can considerably help training in UOGCL. Such results experimentally demonstrate the efficiency of focusing on memory usage rather than minimizing forgetting. We cannot report performances for STAM on Tiny IN since the author did not give corresponding parameters for this dataset and CIFAR100 parameters gave poor performances.

\textbf{Impact of DAA.} To disentangle the impact of DAA compared to standard augmentation, we present in table \ref{tab:avg_acc} the results of our method with $(p,q)=(7,1)$, namely $Ours - (7,1,0,0,0)$ and the results of our method with $(p,q)=(4,1)$ and 1 DAS, 1 DAM, 1 DAC; namely $Ours - (4,1,1,1,1)$. It can be seen that for the same number of augmentations overall, using DAA gives better performances in all considered scenarios.

\textbf{Comparison to supervised methods.} Since very few methods have been designed for UOGCL, we also implemented some typical supervised methods for OGCL. Results displayed in table \ref{tab:avg_acc} show that for small memory sizes, our method can achieve performances close to SCR, a state-of-the-art supervised technique. Specifically, on CIFAR10 with $M=200$, our method performs only $1.5\%$ below SCR. We conjecture that this results from self-supervised methods being less sensitive to overfitting, which is especially important for smaller memory sizes.

\begin{table*}[ht]
\setlength{\tabcolsep}{4pt}
\centering
\vskip 0.03in
\resizebox{\textwidth}{!}{
    \begin{tabular}{l | l | l l | l l | l l l }
    \hline
        \multicolumn{2}{c}{}            & \multicolumn{2}{c}{CIFAR10}  & \multicolumn{2}{c}{CIFAR100} & \multicolumn{3}{c}{Tiny ImageNet} \\
        \hline
        
        \multicolumn{2}{c|}{Method}                       &  \multicolumn{1}{c}{M=200}             &  \multicolumn{1}{c}{M=500}                   &  \multicolumn{1}{c}{M=2k}               &  \multicolumn{1}{c}{M=5k}          &  \multicolumn{1}{c}{M=2k}              &  \multicolumn{1}{c}{M=5k}          &  \multicolumn{1}{c}{M=10k} \\
        \hline\hline
        \multirow{7}{*}{\rotatebox[origin=c]{90}{Supervised}} & offline         & \multicolumn{2}{c|}{86.1$\pm$5.7}                   & \multicolumn{2}{c|}{53.0$\pm$1.8} & \multicolumn{3}{c}{42.3$\pm$3.9}                     \\
                                & fine-tuned      & \multicolumn{2}{c|}{16.6$\pm$2.3}                   & \multicolumn{2}{c|}{3.6$\pm$0.7}                      & \multicolumn{3}{c}{1.4$\pm$0.1}  \\
                    & ER~\cite{rolnick_experience_2019}        & 41.46±3.41 &  52.93±4.39 &  31.37±0.69 &  39.22±1.11 &  11.33±1.17 &  19.4±2.26 &  25.93±3.02   \\
                    & GDUMB~\cite{vedaldi_gdumb_2020}     & 34.06±1.81 &  41.42±1.25 &  15.74±0.61 &  25.53±0.44 &   7.08±0.39 &  13.79±0.76 &  22.35±0.23  \\
                    & SCR~\cite{mai_supervised_2021}       & 49.16±3.02 &  60.28±1.21 &  37.79±0.95 &  47.31±0.34 &  19.76±0.24 &  28.80±0.51 &  34.28±0.28 \\
                    & ER-ACE~\cite{caccia_new_2022}    & 45.25±2.85 &  53.10±2.70 &  33.32±1.14 &  40.60±1.55 &  21.71±0.34 & 27.27±0.95 &  32.57±1.0 \\
                    & GSA~\cite{guo_dealing_2023}       & 52.03±2.14 &  61.30±2.35 &  38.77±1.07 &  48.21±0.99 &  19.35±0.72 &  27.58±0.74 & 34.72±0.82 - \\
        \hline
        \multirow{8}{*}{\rotatebox[origin=c]{90}{Unsupervised}} & STAM      & \multicolumn{2}{c|}{30.54$\pm$0.8}                             & \multicolumn{2}{c|}{8.39$\pm$0.4}                      & \multicolumn{3}{c}{-} \\
                    & SCALE~\cite{yu_scale_2023}     & \multicolumn{2}{c|}{32$\pm\text{1}^\star$}   & \multicolumn{2}{c|}{22$\pm\text{0.1}^\star$}   & \multicolumn{3}{c}{-}   \\
                     & LUMP~\cite{madaan_lump_2022}     & 24.96±1.72 &  25.34±1.06 & 7.42±0.57 &  7.18±0.5 & 4.15±0.5 &  4.55±0.68 &  5.41±0.19  \\
                    & SimSiam-ER~\cite{chen_exploring_2020} & 27.73±1.18 &  30.59±1.21 &  6.91±0.37 &  7.47±0.11 &  5.69±0.32 &  6.49±0.41 &  6.9±0.52 \\
         & BYOL-ER~\cite{grill_bootstrap_2020} & 29.43±0.55 &  29.30±1.01 &  9.39±0.52 &  10.35±0.61 &  5.07±0.39 &  6.19±0.26 &  6.59±0.38 \\
                    & SimCLR-ER~\cite{chen_simple_2020}   & 43.20±2.30   & 48.81±0.78  & 21.2±0.9  & 23.62±0.54  & 12.84±0.7 & 16.7±0.5  & 17.97±0.14 \\
                    & Ours $(7,1,0,0,0)$ & 45.68±2.38 &  52.89±0.57 &  27.27±0.13 &  31.32±0.64 & 13.16±0.37 & 17.9±0.58 &  20.21±0.13  \\
                    & Ours $(4,1,1,1,1)$ & \textbf{48.09±1.22} &  \textbf{56.02±1.34} &  \textbf{29.02±0.77} &  \textbf{33.19±0.9 }&  \textbf{14.79±0.49} &  \textbf{20.35±0.02} &  \textbf{22.06±0.37}  \\
        \hline\hline
    \end{tabular}
}
\caption{Final AA (\%) for all methods on CIFAR10, CIFAR100 and Tiny ImageNet and varying memory sizes $M$. For our method, we reported two set of $(p,q,\#DAM,\#DAC,\#DAS)$ where $\#DAM$, $\#DAC$, $\#DAS$ are the number of DAM, DAC and DAS respectively. Lines corresponding to our method show that 1) using more augmentations can easily improve performances 2) more improvement is achieved using DAA. Each experiment is run 5 times and their average value and standard deviation are reported. The best result and are displayed in bold. Starred values are values reported from the original paper. \label{tab:avg_acc}}
\vskip -0.1in
\end{table*}

\section{Conclusion}
\label{sec:conclusion}
In this paper, we addressed the problem of Unsupervised Online General Continual Learning from the perspective of improving memory usage whereas current state-of-the-art methods propose to cope with catastrophic forgetting. We demonstrated that data augmentation can be enhanced for replay-based methods and proposed a new augmentation strategy, Domain Aware Augmentations, designed for continual learning. We showed the efficiency of focusing on memory usage rather than minimizing
forgetting: with such an approach, we not only surpassed current unsupervised approaches to UOGCL but also narrowed the gap between supervised and unsupervised methods for Online General Continual Learning. Our experiments show that better memory utilization by augmentations implies higher computation costs. As these calculations can be parallelized, the impact on training time remains manageable. Lastly, it should be pointed out that the proposed approach could be adapted to other memory-based methods, with small changes, making it a promising strategy for continual learning.

\bibliography{main}

\begin{thebibliography}{35}
\providecommand{\natexlab}[1]{#1}
\providecommand{\url}[1]{\texttt{#1}}
\expandafter\ifx\csname urlstyle\endcsname\relax
  \providecommand{\doi}[1]{doi: #1}\else
  \providecommand{\doi}{doi: \begingroup \urlstyle{rm}\Url}\fi

\bibitem[Aljundi et~al.(2019{\natexlab{a}})Aljundi, Belilovsky, Tuytelaars, Charlin, Caccia, Lin, and Page-Caccia]{aljundi_online_2019}
Rahaf Aljundi, Eugene Belilovsky, Tinne Tuytelaars, Laurent Charlin, Massimo Caccia, Min Lin, and Lucas Page-Caccia.
\newblock Online {Continual} {Learning} with {Maximal} {Interfered} {Retrieval}.
\newblock In \emph{Advances in {Neural} {Information} {Processing} {Systems}}, volume~32, 2019{\natexlab{a}}.

\bibitem[Aljundi et~al.(2019{\natexlab{b}})Aljundi, Lin, Goujaud, and Bengio]{aljundi_gradient_2019}
Rahaf Aljundi, Min Lin, Baptiste Goujaud, and Yoshua Bengio.
\newblock Gradient based sample selection for online continual learning.
\newblock \emph{Advances in Neural Information Processing Systems}, 32, 2019{\natexlab{b}}.

\bibitem[Buzzega et~al.(2020)Buzzega, Boschini, Porrello, Abati, and Calderara]{buzzega_dark_2020}
Pietro Buzzega, Matteo Boschini, Angelo Porrello, Davide Abati, and Simone Calderara.
\newblock Dark experience for general continual learning: a strong, simple baseline.
\newblock In \emph{Advances in Neural Information Processing Systems}, volume~33, pages 15920--15930, 2020.

\bibitem[Buzzega et~al.(2021)Buzzega, Boschini, Porrello, and Calderara]{buzzega_rethinking_2020}
Pietro Buzzega, Matteo Boschini, Angelo Porrello, and Simone Calderara.
\newblock Rethinking experience replay: a bag of tricks for continual learning.
\newblock In \emph{2020 25th International Conference on Pattern Recognition (ICPR)}, pages 2180--2187. IEEE, 2021.

\bibitem[Caccia et~al.(2022)Caccia, Aljundi, Asadi, Tuytelaars, Pineau, and Belilovsky]{caccia_new_2022}
Lucas Caccia, Rahaf Aljundi, Nader Asadi, Tinne Tuytelaars, Joelle Pineau, and Eugene Belilovsky.
\newblock New insights on reducing abrupt representation change in online continual learning, 2022.

\bibitem[Chawla et~al.(2002)Chawla, Bowyer, Hall, and Kegelmeyer]{chawla_smote_2002}
Nitesh~V. Chawla, Kevin~W. Bowyer, Lawrence~O. Hall, and W.~Philip Kegelmeyer.
\newblock {SMOTE}: synthetic minority over-sampling technique.
\newblock \emph{Journal of Artificial Intelligence Research}, 2002.

\bibitem[Chen et~al.(2020)Chen, Kornblith, Norouzi, and Hinton]{chen_simple_2020}
Ting Chen, Simon Kornblith, Mohammad Norouzi, and Geoffrey Hinton.
\newblock A simple framework for contrastive learning of visual representations.
\newblock In \emph{International conference on machine learning}, pages 1597--1607. PMLR, 2020.

\bibitem[Chen and He(2021)]{chen_exploring_2020}
Xinlei Chen and Kaiming He.
\newblock Exploring simple siamese representation learning.
\newblock In \emph{Proceedings of the IEEE/CVF conference on computer vision and pattern recognition}, pages 15750--15758, 2021.

\bibitem[Davari et~al.(2022)Davari, Asadi, Mudur, Aljundi, and Belilovsky]{davari_probing_2022}
MohammadReza Davari, Nader Asadi, Sudhir Mudur, Rahaf Aljundi, and Eugene Belilovsky.
\newblock Probing representation forgetting in supervised and unsupervised continual learning.
\newblock In \emph{Proceedings of the IEEE/CVF Conference on Computer Vision and Pattern Recognition}, pages 16712--16721, 2022.

\bibitem[Douillard et~al.(2022)Douillard, Ram{\'e}, Couairon, and Cord]{douillard_dytox_2022}
Arthur Douillard, Alexandre Ram{\'e}, Guillaume Couairon, and Matthieu Cord.
\newblock Dytox: Transformers for continual learning with dynamic token expansion.
\newblock In \emph{Proceedings of the IEEE/CVF Conference on Computer Vision and Pattern Recognition}, pages 9285--9295, 2022.

\bibitem[Fini et~al.(2022)Fini, Da~Costa, Alameda-Pineda, Ricci, Alahari, and Mairal]{fini_cassle_2022}
Enrico Fini, Victor G.~Turrisi Da~Costa, Xavier Alameda-Pineda, Elisa Ricci, Karteek Alahari, and Julien Mairal.
\newblock Self-{Supervised} {Models} are {Continual} {Learners}.
\newblock In \emph{2022 {IEEE}/{CVF} {Conference} on {Computer} {Vision} and {Pattern} {Recognition} ({CVPR})}, 2022.

\bibitem[Gatys et~al.(2016)Gatys, Ecker, and Bethge]{gatys_image_2016}
Leon~A. Gatys, Alexander~S. Ecker, and Matthias Bethge.
\newblock Image {Style} {Transfer} {Using} {Convolutional} {Neural} {Networks}.
\newblock In \emph{Computer Vision and Patern Recognition (CVPR)}, 2016.

\bibitem[Grill et~al.(2020)Grill, Strub, Altch{\'e}, Tallec, Richemond, Buchatskaya, Doersch, Avila~Pires, Guo, Gheshlaghi~Azar, et~al.]{grill_bootstrap_2020}
Jean-Bastien Grill, Florian Strub, Florent Altch{\'e}, Corentin Tallec, Pierre Richemond, Elena Buchatskaya, Carl Doersch, Bernardo Avila~Pires, Zhaohan Guo, Mohammad Gheshlaghi~Azar, et~al.
\newblock Bootstrap your own latent-a new approach to self-supervised learning.
\newblock \emph{Advances in neural information processing systems}, 33:\penalty0 21271--21284, 2020.

\bibitem[Guo et~al.(2022)Guo, Liu, and Zhao]{guo_online_2022}
Yiduo Guo, Bing Liu, and Dongyan Zhao.
\newblock Online {Continual} {Learning} through {Mutual} {Information} {Maximization}.
\newblock In \emph{Proceedings of the 39th {International} {Conference} on {Machine} {Learning}}, 2022.

\bibitem[Guo et~al.(2023)Guo, Liu, and Zhao]{guo_dealing_2023}
Yiduo Guo, Bing Liu, and Dongyan Zhao.
\newblock Dealing with cross-task class discrimination in online continual learning.
\newblock In \emph{Proceedings of the IEEE/CVF Conference on Computer Vision and Pattern Recognition}, pages 11878--11887, 2023.

\bibitem[He et~al.(2016)He, Zhang, Ren, and Sun]{he_deep_2015}
Kaiming He, Xiangyu Zhang, Shaoqing Ren, and Jian Sun.
\newblock Deep residual learning for image recognition.
\newblock \emph{Proceedings of the IEEE Conference on Computer Vision and Pattern Recognition}, pages 770--778, 2016.

\bibitem[He et~al.(2020)He, Fan, Wu, Xie, and Girshick]{He_2020_CVPR}
Kaiming He, Haoqi Fan, Yuxin Wu, Saining Xie, and Ross Girshick.
\newblock Momentum contrast for unsupervised visual representation learning.
\newblock In \emph{Proceedings of the IEEE/CVF Conference on Computer Vision and Pattern Recognition (CVPR)}, June 2020.

\bibitem[Hsu et~al.(2018)Hsu, Liu, Ramasamy, and Kira]{hsu_re-evaluating_2019}
Yen-Chang Hsu, Yen-Cheng Liu, Anita Ramasamy, and Zsolt Kira.
\newblock Re-evaluating continual learning scenarios: A categorization and case for strong baselines.
\newblock \emph{arXiv:1810.12488}, 2018.

\bibitem[Huang and Belongie(2017)]{huang_arbitrary_2017}
Xun Huang and Serge Belongie.
\newblock Arbitrary style transfer in real-time with adaptive instance normalization.
\newblock In \emph{Proceedings of the IEEE international conference on computer vision}, pages 1501--1510, 2017.

\bibitem[Khosla et~al.(2020)Khosla, Teterwak, Wang, Sarna, Tian, Isola, Maschinot, Liu, and Krishnan]{khosla_supervised_2020}
Prannay Khosla, Piotr Teterwak, Chen Wang, Aaron Sarna, Yonglong Tian, Phillip Isola, Aaron Maschinot, Ce~Liu, and Dilip Krishnan.
\newblock Supervised contrastive learning.
\newblock \emph{Advances in Neural Information Processing Systems}, 33:\penalty0 18661--18673, 2020.

\bibitem[Krizhevsky et~al.(2009)]{krizhevsky_learning_2009}
Alex Krizhevsky et~al.
\newblock Learning multiple layers of features from tiny images.
\newblock \emph{University of Toronto}, 2009.

\bibitem[Le and Yang(2015)]{le_tiny_2015}
Ya~Le and Xuan Yang.
\newblock Tiny imagenet visual recognition challenge.
\newblock \emph{CS 231N}, 7\penalty0 (7):\penalty0 3, 2015.

\bibitem[Liang et~al.(2023)Liang, Chen, Chen, Ji, and Zhang]{liang_new_2023}
Guoqiang Liang, Zhaojie Chen, Zhaoqiang Chen, Shiyu Ji, and Yanning Zhang.
\newblock New insights on relieving task-recency bias for online class incremental learning.
\newblock \emph{arXiv:2302.08243}, 2023.

\bibitem[Lin et~al.(2014)Lin, Maire, Belongie, Hays, Perona, Ramanan, Doll{\'a}r, and Zitnick]{lin_microsoft_2015}
Tsung-Yi Lin, Michael Maire, Serge Belongie, James Hays, Pietro Perona, Deva Ramanan, Piotr Doll{\'a}r, and C~Lawrence Zitnick.
\newblock Microsoft coco: Common objects in context.
\newblock In \emph{european conference on computer vision (ECCV)}, 2014.

\bibitem[Madaan et~al.(2021)Madaan, Yoon, Li, Liu, and Hwang]{madaan_lump_2022}
Divyam Madaan, Jaehong Yoon, Yuanchun Li, Yunxin Liu, and Sung~Ju Hwang.
\newblock Representational continuity for unsupervised continual learning.
\newblock \emph{arXiv:2110.06976}, 2021.

\bibitem[Mai et~al.(2021)Mai, Li, Kim, and Sanner]{mai_supervised_2021}
Zheda Mai, Ruiwen Li, Hyunwoo Kim, and Scott Sanner.
\newblock Supervised contrastive replay: Revisiting the nearest class mean classifier in online class-incremental continual learning.
\newblock In \emph{Conference on Computer Vision and Pattern Recognition Workshop (CVPRW)}, 2021.

\bibitem[Mai et~al.(2022)Mai, Li, Jeong, Quispe, Kim, and Sanner]{mai_online_2021}
Zheda Mai, Ruiwen Li, Jihwan Jeong, David Quispe, Hyunwoo Kim, and Scott Sanner.
\newblock Online continual learning in image classification: An empirical survey.
\newblock \emph{Neurocomputing}, 2022.

\bibitem[Michel et~al.(2022)Michel, Negrel, Chierchia, and Bercher]{michel_contrastive_2022}
Nicolas Michel, Romain Negrel, Giovanni Chierchia, and Jean-Fmn{\c{c}}ois Bercher.
\newblock Contrastive learning for online semi-supervised general continual learning.
\newblock In \emph{2022 IEEE International Conference on Image Processing (ICIP)}. IEEE, 2022.

\bibitem[Prabhu et~al.(2020)Prabhu, Torr, and Dokania]{vedaldi_gdumb_2020}
Ameya Prabhu, Philip~HS Torr, and Puneet~K Dokania.
\newblock Gdumb: A simple approach that questions our progress in continual learning.
\newblock In \emph{european conference on computer vision (ECCV)}, 2020.

\bibitem[Rolnick et~al.(2019)Rolnick, Ahuja, Schwarz, Lillicrap, and Wayne]{rolnick_experience_2019}
David Rolnick, Arun Ahuja, Jonathan Schwarz, Timothy Lillicrap, and Gregory Wayne.
\newblock Experience {Replay} for {Continual} {Learning}.
\newblock In \emph{Advances in {Neural} {Information} {Processing} {Systems}}, 2019.

\bibitem[Smith et~al.(2019)Smith, Taylor, Baer, and Dovrolis]{smith_unsupervised_2021}
James Smith, Cameron Taylor, Seth Baer, and Constantine Dovrolis.
\newblock Unsupervised progressive learning and the stam architecture.
\newblock \emph{arXiv:1904.02021}, 2019.

\bibitem[Vitter(1985)]{vitter_random_1985}
Jeffrey~S. Vitter.
\newblock Random sampling with a reservoir.
\newblock \emph{ACM Transactions on Mathematical Software}, 1985.

\bibitem[Yu et~al.(2023)Yu, Guo, Gao, and Rosing]{yu_scale_2023}
Xiaofan Yu, Yunhui Guo, Sicun Gao, and Tajana Rosing.
\newblock Scale: Online self-supervised lifelong learning without prior knowledge.
\newblock In \emph{Proceedings of the IEEE/CVF Conference on Computer Vision and Pattern Recognition (CVPR) Workshops}, 2023.

\bibitem[Yun et~al.(2019)Yun, Han, Oh, Chun, Choe, and Yoo]{yun_cutmix_2019}
Sangdoo Yun, Dongyoon Han, Seong~Joon Oh, Sanghyuk Chun, Junsuk Choe, and Youngjoon Yoo.
\newblock Cutmix: Regularization strategy to train strong classifiers with localizable features.
\newblock In \emph{Proceedings of the IEEE/CVF international conference on computer vision}, 2019.

\bibitem[Zhang et~al.(2017)Zhang, Cisse, Dauphin, and Lopez-Paz]{zhang_mixup_2018}
Hongyi Zhang, Moustapha Cisse, Yann~N Dauphin, and David Lopez-Paz.
\newblock mixup: Beyond empirical risk minimization.
\newblock \emph{arXiv:1710.09412}, 2017.

\end{thebibliography}
\end{document}